\documentclass[10pt,twocolumn,letterpaper]{article}

\usepackage{iccv}              % To produce the CAMERA-READY version
\usepackage{multirow}
\usepackage{color}
\usepackage{pifont}

\definecolor{iccvblue}{rgb}{0.21,0.49,0.74}
\usepackage[pagebackref,breaklinks,colorlinks,allcolors=iccvblue]{hyperref}

\def\paperID{7294} % *** Enter the Paper ID here
\def\confName{ICCV}
\def\confYear{2025}

\title{Attention to Trajectory: Trajectory-Aware Open-Vocabulary Tracking}

\author{Yunhao Li$^{1,2}$,  \ \ \ \ Yifan Jiao$^{1,2}$,  \ \ \ \ Dan Meng$^{4}$, \ \ \ \ Heng Fan$^{3,\dagger}$, \ \ \ \ Libo Zhang$^{2,\dagger,*}$ \\
   $^1$Institute of Software Chinese Academy of Science \\
   $^2$University of Chinese Academy of Science \\
   $^3$University of North Texas \ \ \ \ $^4$OPPO Research Institute\\
  \texttt{liyunhao23@mails.ucas.ac.cn, heng.fan@unt.edu, libo@iscas.ac.cn} \\
  $\dagger$Equal Advising \;\;\; $*$Corresponding Author\\
}

\begin{document}
\maketitle
\begin{abstract}
Open-Vocabulary Multi-Object Tracking (OV-MOT) aims to enable approaches to track objects without being limited to a predefined set of categories. Current OV-MOT methods typically rely primarily on instance-level detection and association, often overlooking trajectory information that is unique and essential for object tracking tasks. Utilizing trajectory information can enhance association stability and classification accuracy, especially in cases of occlusion and category ambiguity, thereby improving adaptability to novel classes. Thus motivated, in this paper we propose \textbf{TRACT}, an open-vocabulary tracker that leverages trajectory information to improve both object association and classification in OV-MOT. Specifically, we introduce a \textit{Trajectory Consistency Reinforcement} (\textbf{TCR}) strategy, that benefits tracking performance by improving target identity and category consistency. In addition, we present \textbf{TraCLIP}, a plug-and-play trajectory classification module. It integrates \textit{Trajectory Feature Aggregation} (\textbf{TFA}) and \textit{Trajectory Semantic Enrichment} (\textbf{TSE}) strategies to fully leverage trajectory information from visual and language perspectives for enhancing the classification results. Extensive experiments on OV-TAO show that our TRACT significantly improves tracking performance, highlighting trajectory information as a valuable asset for OV-MOT. Code will be released.
\end{abstract}    
\section{Introduction}

Multi-Object Tracking (MOT) is an important task in computer vision, focusing on the detection and tracking of objects within video sequences. It has many key applications, such as autonomous driving, intelligent surveillance, and robotics. Early MOT research primarily concentrates on a few common categories, \eg, pedestrians and vehicles, and later shifts toward tracking a broader range of categories. Recently, as the demand for practical applications grows, Open-Vocabulary MOT~\cite{li2023ovtrack} is introduced to enable tracking across arbitrary categories, overcoming the limitations imposed by pre-defined tracking categories in training data.

% Multi-Object Tracking (MOT) is a foundational task in computer vision, focusing on the detection and tracking of objects within video sequences. It has broad applications, such as autonomous driving, intelligent surveillance, and robotics. Early MOT researches primarily concentrate on a few common categories, such as pedestrians and vehicles, leading to the establishment of related benchmarks, e.g., MOTChallenge~\cite{dendorfer2021motchallenge}, KITTI~\cite{geiger2013vision}, and DanceTrack~\cite{sun2022dancetrack}. As the demand for practical applications grows, the MOT field shifts toward tracking a broader range of categories, resulting in multi-class datasets like TAO~\cite{dave2020tao} and GMOT-40~\cite{bai2021gmot}. Recently, Li et al.~\cite{li2023ovtrack} make a substantial advancement by challenging the constraints of pre-defined tracking categories set by training data. They introduce Open-Vocabulary MOT (OV-MOT), a framework that leverages Multimodal Large Language Models (MLLMs) to enable tracking across arbitrary categories.

\begin{figure}[t]
 \centering
 \includegraphics[width=0.99\linewidth]{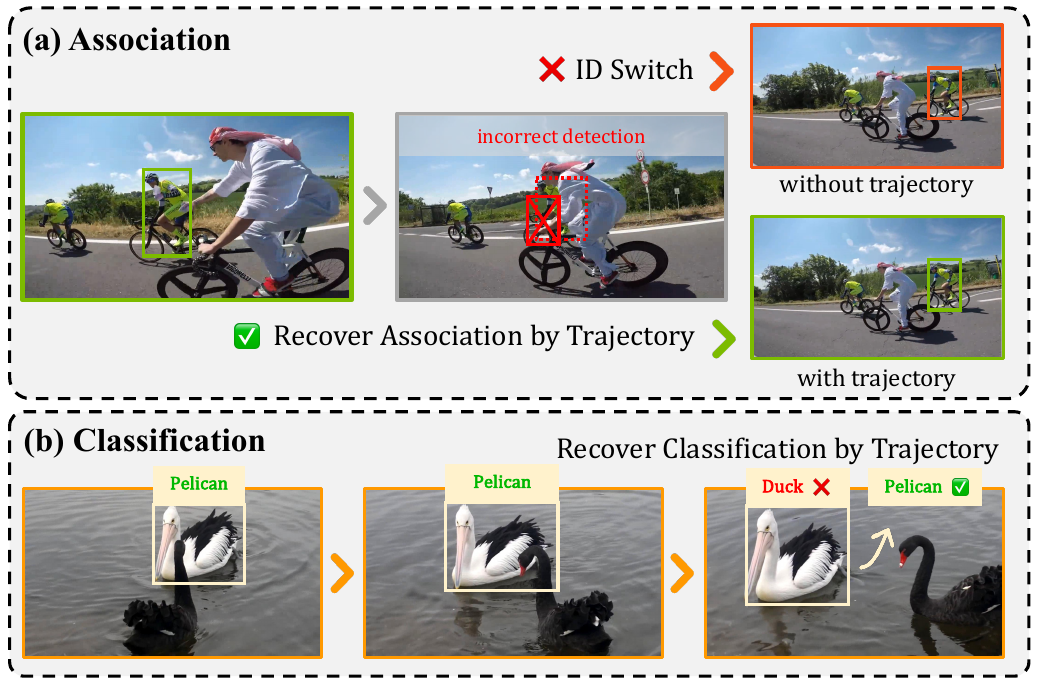}
 %\vspace{-1em}
 \caption{Trajectory information can enhance both association and classification by helping to recover associations disrupted by inaccurate or missed detections (as shown in (a)) and by correcting incorrect classifications (as shown in (b)).}
 \label{fig:intro}
 \vspace{-5mm}
\end{figure}

Despite great advancements, current OV-MOT methods are often constrained by a critical limitation: an overwhelming focus on \textit{instance-level} information, with limited attention to \textit{trajectory-level} insights. Specifically, although recent methods have introduced innovative association strategies for open-vocabulary scenarios, they fail to incorporate trajectory information, which is an important cue in videos and widely utilized in classic MOT approaches. This oversight may prevent current OV-MOT approaches from fully leveraging contextual continuity offered by trajectory\footnote{Please note that, in this paper trajectory information refers to all data related to the trajectory during the tracking process, including its position and classification results from previous frames, among other details.} that is essential to effective tracking, and thus leads to degradation in association and classification (see Fig.~\ref{fig:intro}).

% For instance, OVTrack (CVPR 2023)\cite{li2023ovtrack} employs a diffusion-based data hallucination strategy to generate image pairs for training appearance-based matching. MASA (CVPR 2024)\cite{li2024matching} utilizes the Segment Anything Model (SAM)\cite{kirillov2023segment} to train similar modules on large-scale unannotated image data, and SLAck (ECCV 2024)\cite{li2024slack} further enhances the model’s motion-based matching capabilities.

In this context, we rethink the role of trajectory in OV-MOT and apply it for improvement. Currently, OV-MOT usually contains three steps, including \textit{localization}, \textit{association}, and \textit{classification}. Since the localization mainly depends on the performance of external detectors, it is hard to directly use trajectory information for enhancing localization (see Sec~\ref{sec:discussion} for detailed analysis). However, trajectory information can largely benefit both association and classification, especially in cases with novel classes. For association, the instability of open-vocabulary detection often leads to inaccurate or missed detections in certain frames. In such cases, trajectory information helps recover matches, reducing identity switches (see Fig.~\ref{fig:intro} (a)). For classification, open-vocabulary systems struggle with frequent blurring and occlusion in objects, causing misclassifications. In this situation, trajectory information can aid in distinguishing between incorrect categories accurately (see Fig.~\ref{fig:intro} (b)).

\begin{figure}[t]
 \centering
 \includegraphics[width=0.94\linewidth]{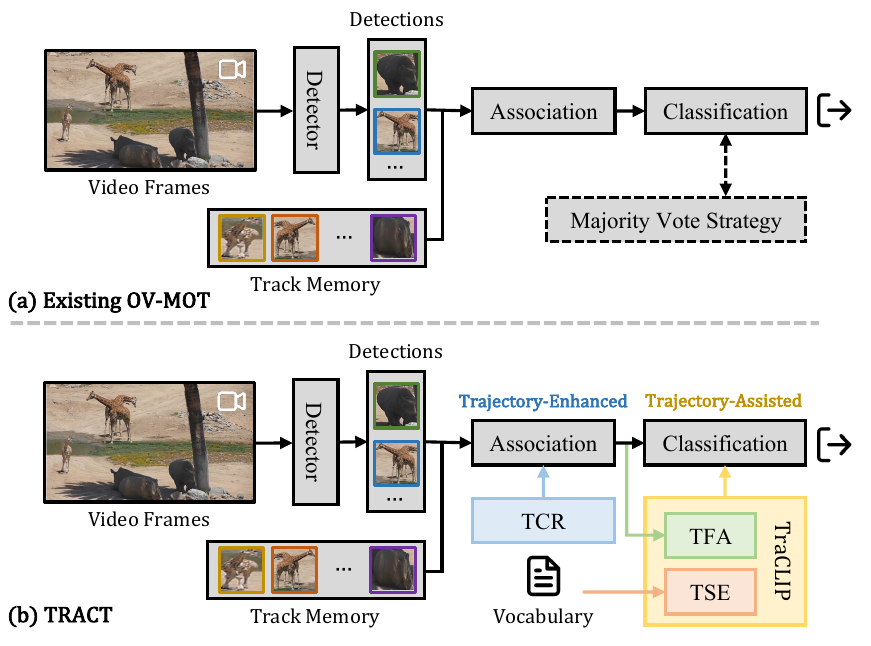}
 \vspace{-1em}
 \caption{Comparison of the overall pipeline between existing OV-MOT approaches and our TRACT. We introduce three strategies, \ie, TCR, TFA, and TSE strategies, to utilize trajectory information in association and classification.}
 \label{fig:comp}
 \vspace{-5mm}
\end{figure}

Motivated by the above, we propose a novel \textit{Trajectory-aware OV-Tracker} (\textbf{TRACT}), a method that comprehensively utilizes trajectory information to improve both object association and classification. We demonstrate the comparison of the overall pipeline between existing OV-MOT approaches and our TRACT in Fig.~\ref{fig:comp}. Built on the top of current mainstream \cite{li2023ovtrack, li2024matching, li2024slack}, it functions as a two-stage tracker that tracks on arbitrary detection results. To better adapt to OV-MOT task, we further divide the tracking stage to association and classification steps (see Section~\ref{sec:pre} for more details). Consequently, the design of TRACT is structured in two key steps: a \textit{Trajectory-Enhanced Association} step and a \textit{Trajectory-Assisted Classification} step. In the initial step, we introduce \textit{Trajectory Consistency Reinforcement} (\textbf{TCR}) strategy, to enhance appearance-based matching models to better capture trajectory dynamics. Specifically, we construct a set of feature banks and category banks to retain memory of previous trajectory information, namely, target visual features and category predictions. Such design strengthens model's ability to maintain \textit{identification} and \textit{category} consistency, thereby aiding in association and indirectly supporting open-vocabulary classification.

On the other hand, in the trajectory assisted classification step we introduce \textbf{TraCLIP}, a plug-and-play method that leverages trajectory information to directly improve classification accuracy. In video sequences, occlusion and blurriness frequently lead to incomplete visual cues, which complicates tracking and especially classification. Trajectory information, however, captures target features under varying occlusion and blur conditions, thus complementing these incomplete visual cues. Therefore, we first propose \textit{Trajectory Feature Aggregation} (\textbf{TFA}) strategy to integrate trajectory features derived from the corresponding detection features. Additionally, since trajectories provide information from multiple viewpoints, trajectory-assisted classification has the potential to offer a more detailed and nuanced understanding of the target compared to image-based classification. In this context, vanilla category names may not be fully or accurately defined, as is typically assumed. We propose \textit{Trajectory Semantic Enrichment} (\textbf{TSE}) strategy, which incorporates attribute-based descriptions as an alternative to relying solely on category names, thereby enriching the semantic context and improving classification precision. With TFA and TSE, TraCLIP leverages the image-text alignment capabilities of CLIP~\cite{radford2021learning} to comprehensively utilize trajectory information for classification.

Thorough experiments on the popular open-vocabulary tracking benchmark OV-TAO~\cite{li2023ovtrack} show the effectiveness of our method, showing satisfactory enhancements in tracking performance in open-vocabulary scenarios. This indicates that trajectory information can effectively contribute to OV-MOT, providing a new research direction. Additionally, this paper aims to encourage researchers to approach the OV-MOT task from a comprehensive video perspective rather than focusing solely on instance-level information.

In summary, in this paper we make the following major contributions: \textbf{(i)} We develop an effective open-vocabulary tracker, termed TRACT, which leverages trajectory-level information to enhance association and classification without bells and whistles; \textbf{(ii)} We propose a plug-and-play trajectory classification method, termed TraCLIP, and introduce the concept of using trajectory itself for classification in OV-MOT; \textbf{(iii)} Extensive experiments demonstrate that our method effectively improves the performance on OV-TAO, in-depth analysis is conducted to provide guidance for future algorithm design.

\section{Related Works}

\subsection{Multi-Object Tracking}

Multi-object tracking (MOT) involves detecting and tracking multiple moving objects in a video sequence while maintaining consistent identities across frames. A popular paradigm in MOT is the “\textit{tracking-by-detection}”. This method~\cite{bewley2016simple, wojke2017simple, du2023strongsort, zhang2022bytetrack, maggiolino2023deep} first performs object detection and then associates detections across frames, forming the basis of many representative methods. In this context, MOT methods often improve their performance by enhancing the detection and matching effectiveness. Another common paradigm is “\textit{joint-detection-and-tracking}”~\cite{yan2022towards, zhang2021fairmot, wang2020towards}, which integrates the tracking and detection into a unified process. Recently, Transformers~\cite{vaswani2017attention} have been introduced into MOT~\cite{zeng2022motr, zhang2023motrv2, gao2023memotr, sun2020transtrack}, significantly surpassing previous trackers in terms of performance.

% Additionally, benchmarks play a key role in advancing the development of MOT. The MOT Challenge dataset~\cite{dendorfer2021motchallenge} covers pedestrian videos in crowd scenes. ImageNet-Vid~\cite{deng2009imagenet} provides annotations for 30 categories across more than 1,000 videos, while TAO~\cite{dave2020tao} extends to 833 object categories, supporting general MOT. Specialized benchmarks such as DanceTrack~\cite{sun2022dancetrack} and SportsMOT~\cite{cui2023sportsmot} target dance and sports scenarios, whereas autonomous driving benchmarks include KITTI~\cite{geiger2013vision} and BDD100K~\cite{yu2020bdd100k}. Moreover, AnimalTrack~\cite{zhang2023animaltrack} focuses on tracking animals in natural environments, while VisDrone~\cite{cao2021visdrone} provides a drone-based tracking benchmark. Additionally, BenSMOT~\cite{li2024beyond} offers supplementary video semantic information, further enriching the available tracking data.

\subsection{Open-Vocabulary Detection}

Open-Vocabulary Detection (OVD) is an emerging task in object detection that aims to identify and localize object categories that are not encountered during the training phase, particularly in few-shot and zero-shot scenarios. In recent years, significant progress has been made in the field of OVD, leading to the proposal of various new algorithms. OVR-CNN~\cite{zareian2021open}, as one of the pioneering works in OVD, successfully applies pretrained vision-language models to detection frameworks, improving recognition capabilities for unseen categories through the integration of image and text. ViLD~\cite{gu2021open} and RegionCLIP~\cite{zhong2022regionclip} utilize the CLIP~\cite{radford2021learning} model, employing knowledge distillation to learn visual region features from classification-oriented models, thus enhancing adaptability in open-world environments. OV-DETR~\cite{zang2022open}, a novel open-vocabulary detector based on the DETR architecture, reformulates the classification task into a binary matching problem between input queries and referent objects to achieve object detection.

\subsection{Open-Vocabulary Multi-Object Tracking}

Open-Vocabulary Multi-Object Tracking (OV-MOT) is a new task in multi-object tracking that aims to identify, locate, and track dynamic objects that are unseen during the training phase. Li et al.~\cite{li2023ovtrack} introduce OVTrack, defining the concept of OV-MOT. They utilize vision-language models for classification and association, enhancing tracking performance through knowledge distillation. Additionally, they employ a data augmentation strategy using a denoising diffusion probabilistic model to learn robust appearance features. They also restructure the TAO dataset~\cite{dave2020tao} into base and novel classes, providing a benchmark for OV-MOT evaluation. Subsequently, they propose MASA~\cite{li2024matching}, which leverages the Segment Anything Model (SAM)~\cite{kirillov2023segment} for object matching. By incorporating unsupervised learning, it automatically generates instance-level correspondences, reducing dependence on annotated data. Recently, the newly proposed SLAack~\cite{li2024slack} employs a unified framework that combines semantic, positional, and appearance information for early-stage association, eliminating the need for complex post-processing heuristics.

% Unlike traditional multi-object tracking methods that rely on a limited set of predefined categories, OV-MOT allows models to track a broader range of categories, offering greater adaptability and flexibility.

\section{Methodology}

\subsection{Preliminary} \label{sec:pre}

In real-world applications, object categories typically follow a long-tailed distribution with a vast vocabulary, reflecting the remarkable diversity that no single dataset can fully encompass. To address this limitation, Li et al.~\cite{li2023ovtrack} introduced Open-Vocabulary MOT, aiming to bridge the gap between conventional MOT and real-world complexity. The mainstream two-stage implementation process of OV-MOT is demonstrated in Fig.~\ref{fig:comp}. For convenient understanding, in this paper we formulate it as follows.

Following the TBD paradigm~\cite{bewley2016simple}, we broadly divide the process into two stages, \ie, detection and tracking. In the first stage, given a video with $N$ frames, a replaceable open-vocabulary detector is first utilized to generate a set of detection results ${\cal R} = \{{\bf b}_i, {\bf c}_i, {\bf f}_i\}^{N}_{i=1}$, where ${\bf b}_i$, ${\bf c}_i$, and ${\bf f}_i$ respectively denotes the set of bounding boxes, category predictions, and extracted target features of the $i^{th}$ frame. 

The second stage is tracking. Unlike conventional MOT, OV-MOT typically involves a highly diverse vocabulary ${\cal V}$ of categories, which presents significant challenges for classification. Consequently, open-vocabulary trackers often perform association in a class-agnostic manner, deferring final classification until the acquisition of the complete trajectory. We define the former as the association step and the latter as the classification step. Notably, although trackers perform association in a class-agnostic manner, the classification prediction for each detection is preserved for later processing. Concretely, trackers obtain a set of trajectories ${\cal T}$ after the association step. Each trajectory ${\bf t} = \{b_i, c_i, f_i\}_{i=1}^n \in {\cal T}$ consists of a series of linked detection results, where $b = [x, y, w, h]$ denotes the 2D bounding box coordinates, $f$ denotes the visual feature, $c$ is the category prediction, and $n$ is the length of ${\bf t}$. Subsequently, in the second step existing trackers utilize the category prediction set $\{c_i\}_{i=1}^n$ to decide the final classification result.

\begin{figure*}[t]
 \centering
 \includegraphics[width=0.93\linewidth]{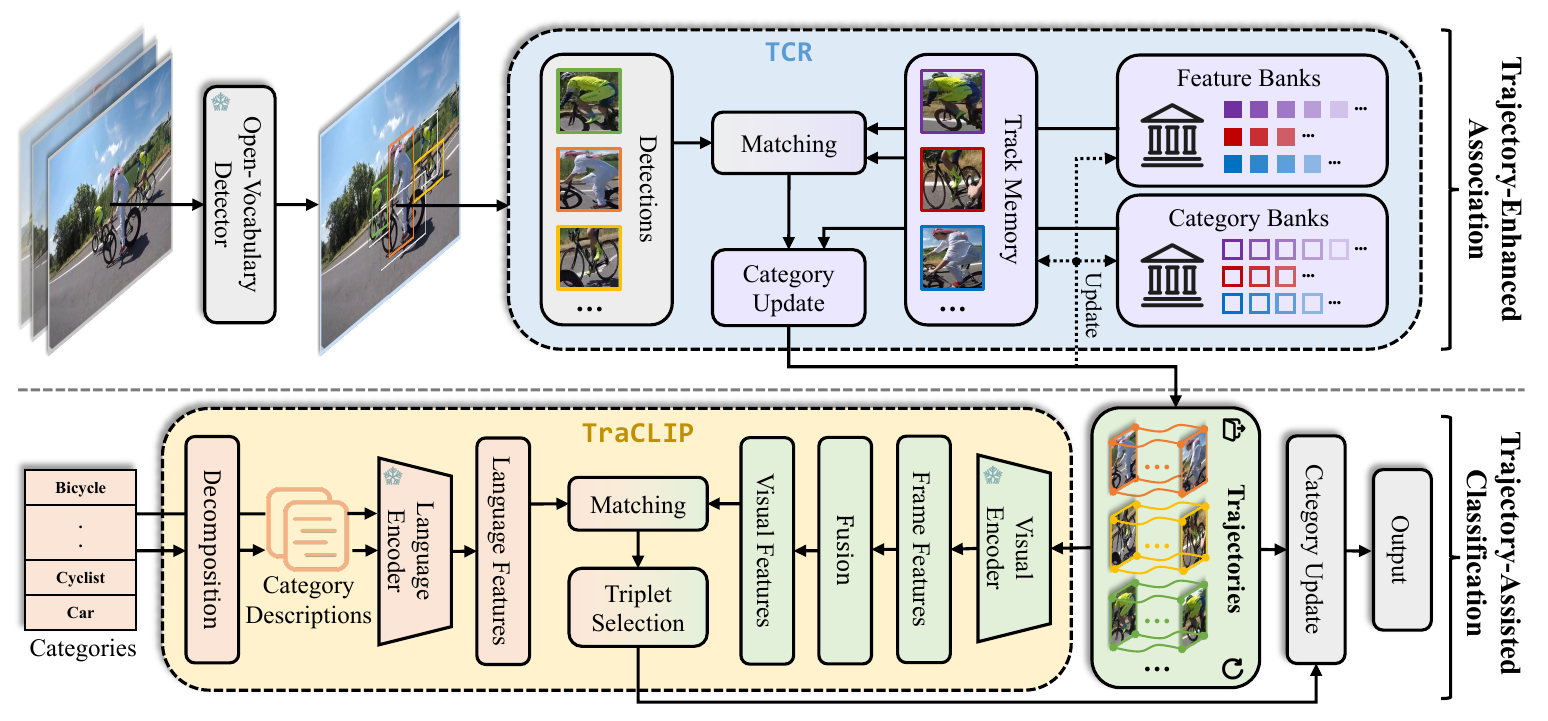}
 %\vspace{-1em}
 \caption{The overall architecture of the proposed TRACT. A replaceable open-vocabulary detector is used to generate boxes of arbitrary categories, and these detection results are used for trajectory association. TRACT leverages trajectory information in both the trajectory-enhanced association and trajectory-assisted classification steps.}
 \label{fig:arch}
 \vspace{-5mm}
\end{figure*}

\subsection{Overview}

%\textbf{TRA}je\textbf{C}tory-aware OV-\textbf{T}racker (TRACT)

In this paper we present the \textit{Trajectory-aware OV-Tracker} (\textbf{TRACT}), to utilize trajectory information in OV-MOT without bells and whistles. As shown in Fig.~\ref{fig:arch}, we address two steps of its design: 1) \textit{Trajectory-Enhanced Association}: we show how to employ trajectory information while associating the detections in Section~\ref{sec:in-asso}. Note that in this step, trajectory information refers to the temporarily stored trajectory segments during the association process. 2) \textit{Trajectory-Assisted Classification}: existing methods determine the category of a trajectory by voting based on the reserved classification results $\{c_i\}_{i=1}^n$. In this paper, we aim to further leverage the trajectory representations $\{f_i\}_{i=1}^n$ to assist obtain the classification results. Therefore, we propose TraCLIP to achieve trajectory-level classification, as described in Section~\ref{sec:post-asso}. Lastly, we introduce the training strategy of TRACT in Section~\ref{sec:training}

\subsection{Trajectory-Enhanced Association} \label{sec:in-asso}

As analyzed in Section~\ref{sec:pre}, in this step all reserved detections are send to association module to obtain object trajectories ${\cal T}$. Based on \cite{li2023ovtrack, li2024matching}, we adopt an appearance-based matching approach as the core association module of TRACT. Please notice, Although OV-MOT methods, including our TRACT, perform class-agnostic association, they retain the classification prediction for each detection within the trajectory (see Section~\ref{sec:pre}). These retained classification predictions are used to determine the final classification result of the trajectory later in the second step. Building upon this, we propose \textit{Trajectory Consistency Reinforcement} (\textbf{TCR}) strategy, a method designed to incorporate trajectory information during association. We decompose its functionality into two aspects:

\vspace{0.3em}
\noindent
\textbf{1) Identification consistency.} For each trajectory alive in the $i^{th}$ frame, we maintain not only a commonly used trajectory memory ${\bf f}$, but also a feature bank ${\bf \bar f} = \{f_{i-j}\}^{n_{\rm bank}}_{j=1}$, recording the target feature embeddings $f$ associated to the trajectory from its previous $n_{\rm bank}$ frames. We update  trajectory memory ${\bf f}$ in a commonly used manner as follows: 
\begin{equation}
\setlength{\abovedisplayskip}{5pt}
\setlength{\belowdisplayskip}{5pt}
    {\bf f}_i = \alpha \times f_i + (1-\alpha) \times {\bf f}_{i-1}
\end{equation}
where ${\bf f}_i$ and $f_i$ represents the trajectory memory and target feature of the target in $i^{th}$ frame, and $\alpha$ is the weighting parameter. We then calculate the similarity between each active trajectory ${\bf t} \in {\cal T}_i$ and each candidate object $r \in {\cal R}_i$:
\begin{equation}
\setlength{\abovedisplayskip}{5pt}
\setlength{\belowdisplayskip}{5pt}
    {\tt S}({\bf t}, r) = \alpha \cdot {\Psi}(f_i, {\bf f}) + (1 - \alpha) \cdot \frac{1}{n_{\rm bank}} \sum_{j=1}^{n_{\rm bank}} {\Psi}(f_i, f_{i-j})
\end{equation}
where $\alpha$ is the weighting parameter, and $f_i \in {\bf f}_i$ denotes the extracted object feature of $r$. We use both cosine similarity and bi-directional softmax for the similarity calculation function ${\Psi}(\cdot)$ as in \cite{li2023ovtrack}. We derive a similarity matrix between each candidate target $r$ and existing trajectories ${\cal T}_i$ , from which we extract the maximum similarity $s_{\rm max}$ and its corresponding trajectory ${\bf t}_{\rm max}$. If $s \ge \tau_{match}$, we assign $r$ to ${\bf t}_{\rm max}$. If $r$ does not have a matching track, we create a new trajectory for $r$ if its confidence score $p_r \ge \tau_{new}$, otherwise we discard it.

\vspace{0.3em}
\noindent
\textbf{2) Category consistency.} As mentioned above, TRACT retains the classification predictions for individual detections during the association process. However, due to the complexity of the OVD task, the classification accuracy achieved by current methods is often suboptimal. Therefore, in TRACT we aim to leverage trajectory information, specific to video-based tasks, to assist this association process. For the $i^{th}$ frame, similar to the approach applied in association, we maintain a category bank ${\bf \bar c} = \{c_{i-j}\}^{n_{\rm clip}}_{j=1}$ for each trajectory to store the category predictions $c$ of the previous $n_{\rm clip}$ frames. When a detected object $r$ with category prediction $c$ is successfully matched to a trajectory ${\bf t}$, we first consider its classification prediction reliable if its confidence $p_r \ge \tau_{\rm high}$. If the confidence falls below $\tau_{\rm high}$ but remains above $\tau_{\rm low}$, we add it to the corresponding category bank. Lastly, if the confidence $p_r < \tau_{\rm low}$, the classification prediction is deemed unreliable. In the first case, the final recorded classification prediction $\bf{c}$ is set as $c$, while in the latter two cases, the classification is determined by a voting mechanism. The process can be depicted as follows:
$$
\bf{c} = \left\{
\begin{array}{lcc}
c & & {\tau_{\rm high} \le p_r \quad \quad \quad} \\
{\tt Vote}({\bf \bar c} \cup \{c\}) & & {\, \tau_{\rm low} \, \le p_r < \tau_{\rm high}} \\
{\tt Vote}({\bf \bar c}) & & {\quad \quad \quad p_r < \tau_{\rm low}}
\end{array}
\right.
$$
where ${\tt Vote}(\cdot)$ stands for the major vote strategy. Note that, the retained classification prediction $\bf{c}$ is not only used for subsequent trajectory classification but also for updating the corresponding category bank.

In summary, during this trajectory-enhanced association step, TRACT predicts object trajectories within a given video and uses TCR to enhance the consistency of identification and classification throughout the association process.

\begin{figure}[t]
 \centering
 \includegraphics[width=0.95\linewidth]{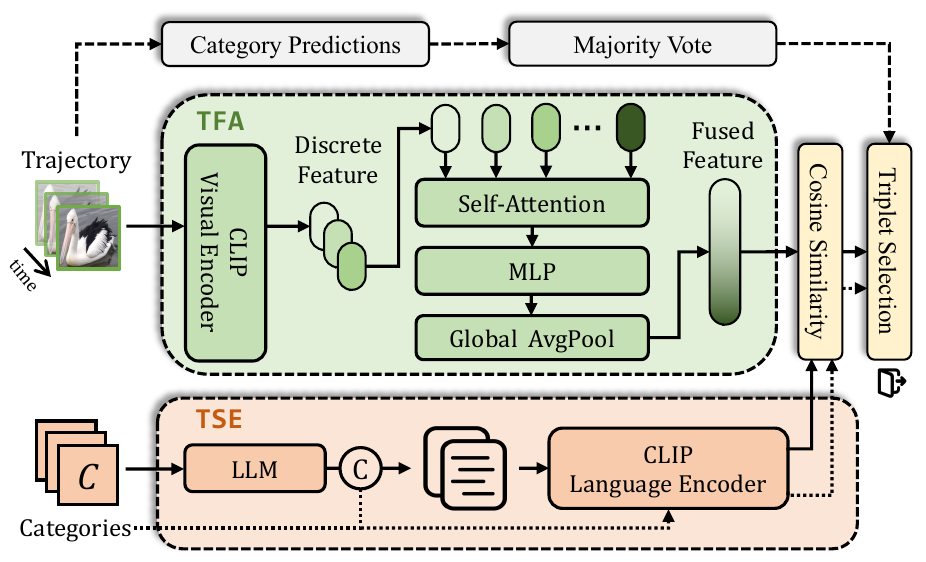}
 %\vspace{-1em}
 \caption{The architecture of the proposed TraCLIP. It approaches both \colorbox[RGB]{255,218,185}{language} and \colorbox[RGB]{184,232,156}{visual} aspects, making full use of trajectory information to assist classification. }
 \label{fig:traclip}
 \vspace{-5mm}
\end{figure}

\subsection{Trajectory-Assisted Classification} \label{sec:post-asso}

Furthermore, we propose a plug-and-play trajectory classification approach, termed \textbf{TraCLIP}, as illustrated in Fig.~\ref{fig:traclip}. Specifically, TraCLIP takes $N$ trajectories ${\cal T}$ and vanilla vocabulary ${\cal V}$ (category names) as input, processes both visual and language information to obtain visual trajectory features and category language features, and then matches them to produce the final trajectory classifications. We address three perspectives of its design:

% ${\cal T} = \{{\cal B}_i, {\cal C}_i\}_{i=1}^N$
\vspace{0.3em}
\noindent
\textbf{1) Visual process.} To leverage the features provided by trajectories under varying occlusion and blur conditions, we introduce \textit{Trajectory Feature Aggregation} (\textbf{TFA}) strategy. Concretely, given a input trajectory ${\bf t} \in {\cal T}$, we first sample them based on the detection confidence, obtaining a sample clip ${\bf \dot t}$ of length $n_{\rm clip}$. If the trajectory length is already less than $n_{\rm clip}$, no sampling is performed. We then use the CLIP visual encoder to extract its 2D feature ${\bf \dot f} \in {\mathbb R}^{{\rm n} \times d}$ frame by frame, where ${\rm n}$ is the length of the sample and feature dimension $d = 768$. We consider ${\bf \dot f}$ as a sequential data, and use self-attention and MLP to get self-enhanced feature ${\bf \tilde f}$:
\begin{align}
\setlength{\abovedisplayskip}{5pt}
\setlength{\belowdisplayskip}{5pt}
    {\bf \ddot f} = {\bf \dot f} + {\tt SA}({\tt LN}({\bf \dot f})) \\
    {\bf \tilde f} = {\bf \ddot f} + {\tt MLP}({\tt LN}({\bf \ddot f}))
\end{align}
where ${\tt SA}({\bf x})$ denotes self-attention with ${\bf x}$ generating query, key, and value as in \cite{vaswani2017attention}, ${\tt MLP}(\cdot)$ denotes the multi-layer perception, and ${\tt LN}(\cdot)$ is a layer normalization function. Finally, we generate the fused trajectory feature by global average pooling ${\bf f}^{\it traj} = \{{\tt AvgPool}({\bf \tilde f}_i)\}_{i=1}^{\rm n}$. We have explored additional fusion methods, please kindly refer to the \textbf{supplementary material} due to limited space.

\vspace{0.3em}
\noindent
\textbf{2) Language process.} Since trajectories provide richer feature information, \eg, target characteristics from different perspectives and lighting conditions, relying solely on category names often results in incomplete language features. Therefore, to fully utilize trajectory information, we introduce \textit{Trajectory Semantic Enrichment} (\textbf{TSE}) strategy to enhance semantics using attribute information. Given the input vanilla vocabulary ${\cal V} = {{\cal C}^{{\rm base}} \cup {\cal C}^{{\rm novel}}}$, we use Large Language Models (LLMs) to decouple them into various attribute descriptions (see Fig.\ref{fig:traclip}). Specially, to better employ LLMs to enrich category contexts, we carefully design a prompt template to ensure accurate decomposition, \ie, ``\textit{Provide a brief description of the \{{\rm category}\} focusing on two to three visual attributes}''. In this work we prompt ChatGPT to generate attribute answers, and then concatenate them with the corresponding category as follows: 
\begin{equation}
    {\cal A} = {\tt Concat}({\cal V}, \Phi({\cal V}))
\end{equation}
where $\Phi({\cdot})$ denotes LLM processing operation. Please refer to \textbf{supplementary material} for more details. With the enriched category texts ${\cal A}$ available, we use the CLIP language encoder to extract two sets of category language features:
\begin{align}
    {\cal F}^{\it attr} = {\tt Linear}({\tt Enc}({\cal A})) \\
    {\cal F}^{\it cate} = {\tt Linear}({\tt Enc}({\cal V}))
\end{align}
where ${\tt Enc}({\cdot})$ represents the CLIP language encoder, and ${\tt Linear}(\cdot)$ stands for a linear projection layer. ${\cal F}^{\it attr}$ and ${\cal F}^{\it cate}$ represent the attribute-assisted language feature and the vanilla language feature, respectively.

\begin{table*}[!t]\scriptsize
    \centering
    \renewcommand{\arraystretch}{1.01}
    \tabcolsep=2mm
    \caption{Comparison with state-of-the-art trackers on OV-TAO dataset. The experiments are grouped based on different detectors, which we consider to be more reasonable. The best and second best results within each detection setting are highlighted in \textbf{bold} and \underline{underline}.}\vspace{-1mm}
    \resizebox{0.95\textwidth}{!}{
    \begin{tabular}{c|r|c|cccc|cccc}
        \toprule[1.2pt]
        \multirow{1}{*}{\textbf{Detector}} & \multicolumn{2}{c|}{\textbf{Method}} & \multicolumn{4}{c|}{\textbf{Base}} & \multicolumn{4}{c}{\textbf{Novel}} \\
        \midrule[0.8pt]
        \multicolumn{2}{c|}{\textbf{Validation Set}} & \textbf{Publication} & \textbf{TETA}$\uparrow$ & \textbf{LocA}$\uparrow$ & \textbf{AssA}$\uparrow$ & \textbf{ClsA}$\uparrow$ & \textbf{TETA}$\uparrow$ & \textbf{LocA}$\uparrow$ & \textbf{AssA}$\uparrow$ & \textbf{ClsA}$\uparrow$ \\
        \midrule[0.6pt]
        % ViLD
        \multirow{5}{*}{ViLD~\cite{gu2021open}} & DeepSORT~\cite{wojke2017simple} & ICIP 2017 & 26.9 & 47.1 & 15.8 & 17.1 & 21.1 & 46.4 & 14.7 & \underline{2.3} \\
        & Tracktor++~\cite{bergmann2019tracking} & ICCV 2019 & 28.3 & 47.4 & 20.5 & 17.0 & 22.7 & 46.7 & 19.3 & 2.2 \\
        & OVTrack~\cite{li2023ovtrack} & CVPR 2023 & 35.5 & 49.3 & 36.9 & \underline{20.2} & 27.8 & 48.8 & 33.6 & 1.5 \\
        & MASA~\cite{li2024matching} & CVPR 2024 & \underline{37.5} & \textbf{55.2} & \underline{37.9} & 19.3 & \underline{30.3} & \textbf{52.8} & \underline{35.9} & \underline{2.3} \\
        & TRACT & Ours & \textbf{38.5} & \underline{55.0} & \textbf{39.0} & \textbf{21.5} & \textbf{31.3} & \underline{52.7} & \textbf{37.8} & \textbf{3.4} \\
        \midrule[0.6pt]
        % RegionCLIP
        \multirow{5}{*}{RegionCLIP~\cite{zhong2022regionclip}} & DeepSORT~\cite{wojke2017simple} & ICIP 2017 & 28.4 & 52.5 & 15.6 & 17.0 & 24.5 & 49.2 & 15.3 & 9.0 \\
        & Tracktor++~\cite{bergmann2019tracking} & ICCV 2019 & 29.6 & 52.4 & 19.6 & 16.9 & 25.7 & 50.1 & 18.9 & 8.1 \\
        & ByteTrack~\cite{zhang2022bytetrack} & ICCV 2019 & 29.4 & 52.3 & 19.8 & 16.0 & 26.5 & 50.8 & 20.9 & 8.0 \\
        & OVTrack~\cite{li2023ovtrack} & CVPR 2023 & 36.3 & 53.9 & 36.3 & \underline{18.7} & 32.0 & 51.4 & 33.2 & 11.4 \\
        & MASA~\cite{li2024matching} & CVPR 2024 & \underline{36.7} & \textbf{54.4} & \underline{38.5} & 17.3 & \underline{33.6} & \underline{53.7} & \underline{35.3} & \underline{11.8} \\
        & TRACT & Ours & \textbf{37.9} & \underline{54.2} & \textbf{39.4} & \textbf{20.2} & \textbf{34.4} & \textbf{54.0} & \textbf{36.0} & \textbf{13.3} \\
        \midrule[0.6pt]
        \multirow{5}{*}{YOLO-World~\cite{cheng2024yolo}} & DeepSORT~\cite{wojke2017simple} & ICIP 2017 & 27.3 & 47.1 & 16.5 & 17.9 & 21.5 & 48.9 & 14.9 & 3.8 \\
        & ByteTrack~\cite{zhang2022bytetrack} & ECCV 2022 & 28.5 & 46.8 & 19.2 & 17.1 & 22.9 & 50.1 & 19.7 & 3.3 \\
        & OC-SORT~\cite{cao2023observation} & CVPR 2023 & 31.2 & \underline{51.0} & 18.8 & 16.9 & 24.4 & 53.3 & 20.3 & 3.7 \\
        & MASA~\cite{li2024matching}& CVPR 2024 & \underline{38.2} & \textbf{54.9} & \textbf{41.0} & \underline{18.6} & \underline{32.2} & \underline{55.2} & \underline{37.9} & \underline{4.4} \\
        & TRACT & Ours & \textbf{39.4} & \textbf{54.9} & \underline{40.6} & \textbf{22.6} & \textbf{33.7} & \textbf{56.0} & \textbf{39.8} & \textbf{5.3} \\
        \bottomrule[0.8pt]
        \toprule[0.8pt]
        % 测试集结果
        \multicolumn{2}{c|}{\textbf{Test Set}} & \textbf{Publication} & \textbf{TETA}$\uparrow$ & \textbf{LocA}$\uparrow$ & \textbf{AssA}$\uparrow$ & \textbf{ClsA}$\uparrow$ & \textbf{TETA}$\uparrow$ & \textbf{LocA}$\uparrow$ & \textbf{AssA}$\uparrow$ & \textbf{ClsA}$\uparrow$ \\
        \midrule[0.6pt]
        \multirow{5}{*}{ViLD~\cite{gu2021open}} & DeepSORT~\cite{wojke2017simple} & ICIP 2017 & 24.5 & 43.8 & 14.6 & 15.2 & 17.2 & 38.4 & 11.6 & 1.7 \\
        & Tracktor++~\cite{bergmann2019tracking} & ICCV 2019 & 26.0 & 44.1 & 19.0 & 14.8 & 18.0 & 39.0 & 13.4 & 1.7 \\
        % & ByteTrack~\cite{zhang2022bytetrack} & ECCV 2022 & & & & & & & & \\
        & OVTrack~\cite{li2023ovtrack} & CVPR 2023 & 32.6 & 45.6 & 35.4 & \underline{16.9} & 24.1 & 41.8 & 28.7 & \underline{1.8} \\
        & MASA~\cite{li2024matching} & CVPR 2024 & \underline{35.2} & \textbf{52.5} & \underline{37.9} & 15.3 & \underline{26.6} & \underline{47.9} & \underline{30.6} & 1.3 \\
        & TRACT & Ours & \textbf{36.2} & \underline{52.3} & \textbf{39.1} & \textbf{17.2} & \textbf{27.3} & \textbf{48.2} & \textbf{30.7} & \textbf{3.1} \\
        \midrule[0.6pt]
        \multirow{5}{*}{RegionCLIP~\cite{zhong2022regionclip}} & DeepSORT~\cite{wojke2017simple} & ICIP 2017 & 27.0 & 49.8 & 15.1 & 16.1 & 18.7 & 41.8 & 9.1 & 5.2 \\
        & Tracktor++~\cite{bergmann2019tracking} & ICCV 2019 & 28.0 & 49.4 & 18.8 & 15.7 & 20.0 & 42.4 & 12.0 & 5.7 \\
        & ByteTrack~\cite{zhang2022bytetrack} & ECCV 2022 & 28.7 & 51.5 & 19.9 & 14.5 & 20.4 & 43.0 & 13.5 & 4.9 \\
        & OVTrack~\cite{li2023ovtrack} & CVPR 2023 & 34.8 & 51.1 & 36.1 & \underline{17.3} & 25.7 & \underline{44.8} & 26.2 & 6.1 \\
        & MASA~\cite{li2024matching} & CVPR 2024 & \underline{36.5} & \textbf{53.2} & \underline{39.0} & \underline{17.3} & \underline{26.}8 & \underline{44.8} & \underline{29.5} & \underline{6.2} \\
        & TRACT & Ours & \textbf{37.3} & \underline{53.0} & \textbf{39.4} & \textbf{19.3} & \textbf{28.8} & \textbf{45.3} & \textbf{30.1} & \textbf{10.8} \\
        \midrule[0.6pt]
        \multirow{5}{*}{YOLO-World~\cite{cheng2024yolo}} & DeepSORT~\cite{wojke2017simple} & ICIP 2017 & 25.1 & 43.3 & 15.6 & 13.0 & 16.9 & 40.5 & 11.8 & 8.8 \\
        & ByteTrack~\cite{zhang2022bytetrack} & ICCV 2019 & 26.6 & 44.1 & 19.3 & 11.7 & 18.4 & 41.3 & 15.1 & 5.0 \\
        & OC-SORT~\cite{cao2023observation} & CVPR 2023 & 28.9 & 49.0 & 19.1 & 9.9 & 20.6 & 48.3 & 14.8 & 5.8 \\
        & MASA~\cite{li2024matching} & CVPR 2024 & \underline{34.9} & \textbf{51.8} & \underline{39.7} & \underline{13.2} & \underline{32.2} & \underline{51.4} & \textbf{36.2} & \underline{9.2} \\
        & TRACT & Ours & \textbf{36.1} & \underline{51.6} & \textbf{40.7} & \textbf{15.9} & \textbf{33.3} & \textbf{51.8} & \underline{35.9} & \textbf{12.0} \\
        \bottomrule[1.2pt]
    \end{tabular}
    \label{tab:sota-comparison}\vspace{-3mm}
    }
\end{table*}

\vspace{0.3em}
\noindent
\textbf{3) Triplet selection.} At this point, we have obtained two sets of language features ${\cal F}^{\it cate}$, ${\cal F}^{\it attr}$ and a set of visual features ${\cal F}^{\it traj} = \{{\bf f}^{\it traj}_i\}_{i=1}^n$ representing each trajectory, where $n$ denotes the length of the trajectory. Together with the classification predictions $\{c_i\}_{i=1}^n$ retained in the association step, for each trajectory ${\bf t}$ we obtain three classification results along with the corresponding similarity scores. Specifically, we first compute the affinity between its visual features and two types of language features, as follows:
\begin{equation}
{\cal Z}({\bf t}) = [{\tt Cos}({\bf f}_{\bf t}, {\cal F}_1^*), {\tt Cos}({\bf f}_{\bf t}, {\cal F}_2^*), \cdot\cdot\cdot\ , {\tt Cos}({\bf f}_{\bf t}, {\cal F}^*_{|{\cal V}|})]
\end{equation}
where ${\tt Cos}({\bf x}, {\bf y}) = \frac{\mathbf{x} \cdot \mathbf{y}}{\|\mathbf{x}\| \|\mathbf{y}\|}$ represents the cosine similarity, ${\bf f}_{\bf t} \in {\cal F}^{\it traj}$ is the visual feature of ${\bf t}$, and ${\cal F}_i^*$ denotes the $i^{th}$ language feature in ${\cal F}^{\it cate}$ or ${\cal F}^{\it attr}$. We select the classification with the highest similarity score, yielding two classification results ${\bf v}_{\rm cate}$ and ${\bf v}_{\rm attr}$ along with their similarity scores ${\bf s}_{\rm cate}$ and ${\bf s}_{\rm attr}$. Furthermore, we apply a majority vote strategy to obtain the third classification result, represented as ${\bf v}_{\rm det} = {\tt Vote}(c_1, c_2, \dots, c_{|{\cal V}|})$, and use its proportion as the similarity score ${\bf s}_{\rm det}$. Finally, the result with the \textit{highest} similarity score is selected as the final output.
% \begin{equation}
% \mathbf{v}_{\mathrm{final}} = \arg\max_{\mathbf{v} \in \{ \mathbf{v}_{\mathrm{cate}},\, \mathbf{v}_{\mathrm{attr}},\mathbf{v}_{\mathrm{det}} \}} \Bigl\{ \, s(\mathbf{v}_{\mathrm{cate}}),\, s(\mathbf{v}_{\mathrm{attr}}),\, s(\mathbf{v}_{\mathrm{det}}) \, \Bigr\}
% \end{equation}
% where ${\bf v}_{\rm final}$ is the final classification output.

\subsection{Training Strategy} \label{sec:training}

In the trajectory-enhanced association step, both of our proposed trajectory banks are training-free, so rather than designing a specific training method, we turn to the general training approach of appearance-based matching models. In specific, we adopt the training approach from \cite{li2024matching} and employ a contrastive learning method.

% On the other hand, for the trajectory-assisted classification step, we initialize TraCLIP using CLIP~\cite{radford2021learning} weights with ViT-L/14 as backbone, and we freeze both the language encoder and visual encoder during training. Meanwhile, we follow CLIP to use the contrastive learning loss. For training data, we sequentially utilized the LVIS~\cite{gupta2019lvis}, YouTube-VIS~\cite{Yang2019vis}, and TAO datasets~\cite{dave2020tao}. Concretely, we use the target trajectories and category names from these datasets as input data and labels. Since LVIS is an image dataset rather than a video dataset, we apply ${n_{\rm clip}}$ distinct data augmentation techniques to each target to generate trajectory data. These techniques included, \eg, random rotation, random erasure, and random scaling. Please kindly refer to the \textbf{supplementary material} for more details.

For the trajectory-assisted classification, we initialize TraCLIP with CLIP~\cite{radford2021learning} weights using ViT-L/14 as the backbone, freezing both the language and visual encoders during training. We adopt CLIP’s contrastive loss and use LVIS~\cite{gupta2019lvis}, YouTube-VIS~\cite{Yang2019vis}, and TAO~\cite{dave2020tao} training set as training data. Specifically, target trajectories and category names from these datasets serve as input and labels. Since LVIS is an image dataset, we generate trajectory data for each target with ${n_{\rm clip}}$ augmentations, such as random rotation, erasure, and scaling. Note that, during the entire training process, we only used \textit{known} object categories. Please refer to the \textbf{supplementary material} for more details.

\section{Experiments}

\subsection{Experimental Setup} \label{sec:setup}

\noindent
\textbf{Benchmark.} We conduct experiments on the large-scale open-vocabulary dataset OV-TAO, extended from TAO~\cite{dave2020tao}, which includes 2,907 sequences and over 800 categories. OV-TAO follows the  classification scheme inLVIS~\cite{gupta2019lvis} by dividing categories into \textit{base} (common) and \textit{novel} (rare) classes. This setup mirrors the real-world scenarios and can reflect the  adaptability trackers in handling rare categories.

\vspace{0.3em}
\noindent
\textbf{Metrics.} Following~\cite{li2023ovtrack, li2024matching, li2024slack}, we use Tracking-Every-Thing Accuracy (TETA) metric~\cite{li2022tracking}, which disentangles MOT evaluation into three subfactors: Localization Accuracy (LocA), Association Accuracy (AssocA) and Classification Accuracy (ClsA). Please note, all our experiments are conducted under the open-vocabulary setting.

\vspace{0.3em}
\noindent
\textbf{Implementation details.} We conduct experiments with 4 Nvidia Tesla V100 GPUs. We set the batch size to 256 per GPU and use the AdamW optimizer to train the model. Please note again that \textit{only} ${\cal C}^{{\rm base}}$ categories are used in training. The initial learning rate is set to $1\times10^{-4}$ , and the weight decay is set to $1\times10^{-5}$. During inference, we set $n_{\rm bank} = 15$ in the association step and $n_{\rm clip} = 5$ in the classification step. See \textbf{supplementary material} for details.

\begin{figure}[t]
 \centering
 \includegraphics[width=0.95\linewidth]{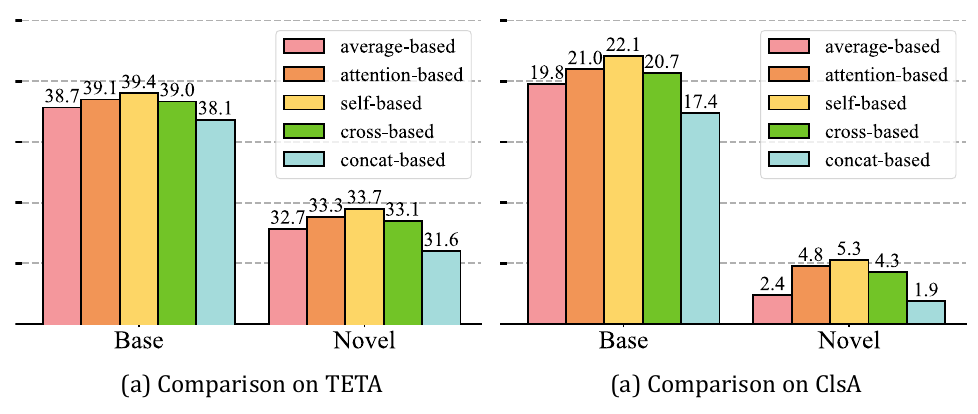}
 %\vspace{-1em}
 \caption{Comparison of different fusion mechanisms on the validation set of OV-TAO~\cite{li2023ovtrack}, using TETA (a) and ClsA (b) metrics.}
 \label{fig:fusion-analysis}
 % \vspace{-5mm}
\end{figure}

\begin{table}[!t]\scriptsize
    \centering
    \renewcommand{\arraystretch}{1}
    \tabcolsep=2.5mm
    \caption{Ablation studies to evaluate the contribution of the proposed strategies in TRACT. The best results are in \textbf{bold}.}\vspace{-1mm}
    \resizebox{0.45\textwidth}{!}{
    \begin{tabular}{ccccccc}
        \toprule[1.2pt]
        TCR & TFA & TSE & \textbf{TETA} & \textbf{LocA} & \textbf{AssA} & \textbf{ClsA}\\
        \midrule[0.8pt]
        & & & 37.5 & \textbf{55.1} & 40.1 & 16.9 \\
        \ding{51} & & & 37.6 & 55.0 & \textbf{40.6} & 17.3 \\
        \ding{51} & \ding{51} & & 38.5 & 54.9 & 40.5 & 19.9 \\
        & \ding{51} & & 38.4 & 54.9 & 40.4 & 19.8 \\
        & \ding{51} & \ding{51} & 38.4 & 54.9 & 40.4 & 19.7 \\
        \ding{51} & \ding{51} & \ding{51} & \textbf{38.6} & 54.9 & \textbf{40.6} & \textbf{20.3} \\
        \bottomrule[1.2pt]
    \end{tabular}
    \label{tab:strategies-ablation}\vspace{-5mm}
    }
\end{table}

\begin{table}[!t]
    % \vspace {-2.0em}
    \centering
    \renewcommand{\arraystretch}{1}
    \tabcolsep=1.0mm
    \caption{Comparison experiments of TSE effect.}\vspace{-1mm}
    \resizebox{0.9\linewidth}{!}{
    \begin{tabular}{c|cc|c|cc}
        \toprule[1.2pt]
         &\textbf{ ClsA (\textit{base})} & \textbf{ClsA (\textit{novel})} & & \textbf{ClsA (\textit{base}}) & \textbf{ClsA (\textit{novel})} \\
        \midrule[0.8pt]
        with TSE & 20.2 & 13.3 & w/o TSE & 21.6 & 10.9  \\
        \bottomrule[1.2pt]
    \end{tabular}
    \label{tab:tse}
    }
\end{table}

\subsection{Comparison to State-of-the-Art}

We conduct experiments on both validation and test sets of TAO. Please note, considering the strong correlation between current OV-MOT and OVD methods, we group the experimental results based on the different OVD models used to ensure fairness in comparison. Concretely, we first use two typically used~\cite{li2023ovtrack, li2024matching} detector ViLD~\cite{gu2021open} and RegionCLIP~\cite{zhong2022regionclip}, and then explore a state-of-the-art OVD method, YOLO-World~\cite{cheng2024yolo}. For ViLD and RegionCLIP, we use the same detection results as in OVTrack~\cite{li2023ovtrack}, while for YOLO-World, we utilize the officially provided weights. Given the limited data volume and incomplete annotations~\cite{li2022tracking} in the TAO training set, we refrain from further fine-tuning. Throughout the comparison, we focus primarily on comparisons within each group.

% As depicted in Tab.~\ref{tab:sota-comparison}, our TRACT consistently achieves top-tier performances across nearly all metrics, demonstrating impressive results using various detectors. For instance, when using the state-of-the-art open-vocabulary detector YOLO-World, it achieves 39.4$\%$ / 33.7$\%$, for base/novel classes respectively, in TETA on the validation set, and 35.7$\%$ / 33.1$\%$ (base/novel classes) on the test set. Particularly, TRACT shows impressive improvements on the ClsA metric. For example, comparing to the currently best tracker MASA~\cite{li2024matching}, TRACT achieves improvements of +2.0$\%$ / +4.6$\%$ (for base/novel classes) on the test set using RegionCLIP, and gains +1.9$\%$ / 1.5$\%$ on the validation set. These results suggest that incorporating trajectory information in OV-MOT is beneficial and holds significant potential.

As shown in Tab.\ref{tab:sota-comparison}, TRACT consistently achieves top-tier results on nearly all metrics, demonstrating strong results with various detectors. For instance, when using the state-of-the-art open-vocabulary detector YOLO-World, it achieves TETA scores of 39.4$\%$ and 33.7$\%$ for base and novel classes, respectively, on the validation set, and 35.7$\%$ and 33.1$\%$ on the test set. Notably, TRACT demonstrates significant improvements on the ClsA metric. Compared to the current leading tracker MASA\cite{li2024matching}, TRACT shows gains of +2.0$\%$ and +4.6$\%$ (base/novel classes) on the test set with RegionCLIP and +1.9$\%$ and +1.5$\%$ on the validation set. These results indicate that incorporating trajectory information in OV-MOT is both beneficial and promising.

Due to limited space, we provide the visualization results in the\textbf{ supplementary material} to show the effectiveness of TRACT and its superiority in handling object occlusion.

% TRACT achieves improvements of +2.2$\%$/+2.9$\%$/+4.0$\%$ on base classes using three distinct detectors on the validation set, and gains +1.1$\%$/+1.5$\%$/+0.9$\%$ on novel classes. On the test set, it also achieves +1.9$\%$/+2.0$\%$/+2.7$\%$ for base classes and +1.8$\%$/+4.6$\%$/+2.8$\%$ for novel classes.

% This validates the effectiveness of our proposed strategies and the trajectory-aware approach by achieving superior performance compared to state-of-the-art MOT and OV-MOT methods.

% \begin{table}[!t]\scriptsize
%     \centering
%     \renewcommand{\arraystretch}{1}
%     \tabcolsep=3mm
%     \caption{Comparison with state-of-the-art trackers on the newly proposed OVT-B dataset. The best results are highlighted in \textbf{bold}.}
%     \resizebox{0.45\textwidth}{!}{
%     \begin{tabular}{c|cccc}
%         \toprule[1.2pt]
%         \textbf{Method} & \textbf{TETA} & \textbf{LocA} & \textbf{AssA} & \textbf{ClsA} \\
%         \midrule[0.8pt]
%         ByteTrack~\cite{zhang2022bytetrack} & 20.1 & 36.1 & 12.4 & 11.9 \\
%         OC-SORT~\cite{cao2023observation} & 16.0 & 31.2 & 4.3 & 12.3 \\
%         StrongSORT~\cite{du2023strongsort} & 24.8 & 60.8 & 66.1 & 12.2 \\
%         OVTrack~\cite{li2023ovtrack} & 46.1 & 60.8 & 66.1 & 11.5 \\
%         OVTrack+~\cite{liang2024ovt} & 47.0 & 62.0 & 67.7 & 11.3 \\
%         \bottomrule[1.2pt]
%     \end{tabular}
%     \label{tab:ovt-comparison}
%     }
% \end{table}

\begin{table}[!t]\scriptsize
    \centering
    \renewcommand{\arraystretch}{1}
    \tabcolsep=1.5mm
    \caption{Ablation studies of $n_{\rm bank}$ on the validation set of OV-TAO . Please note, here we do not use TraCLIP to ensure a clear comparison. The best results are highlighted in \textbf{bold}. We use the average time per sequence to measure the model speed.}\vspace{-1mm}
    \resizebox{0.4\textwidth}{!}{
    \begin{tabular}{c|ccccc}
        \toprule[1.2pt]
        $n_{\rm bank}$ & \textbf{TETA} & \textbf{LocA} & \textbf{AssA} & \textbf{ClsA} & \textbf{Speed}(\textit{s/seq}) $\downarrow$ \\
        \midrule[0.8pt]
        5 & 37.56 & \textbf{55.04} & 41.12 & 16.52 & 1.52 \\
        10 & 37.54 & 55.01 & 40.64 & 16.96 & 1.56 \\
        15 & \textbf{37.62} & \textbf{55.04} & 40.58 & \textbf{17.27} & 1.59 \\
        20 & 37.51 & 55.03 & 40.53 & 16.97 & 1.64 \\
        25 & 37.61 & 55.01 & \textbf{40.74} & 17.09 & 1.70 \\
        \bottomrule[1.2pt]
    \end{tabular}
    \label{tab:bank-ablation}
    }
\end{table}

\subsection{Analysis on TraCLIP} \label{sec:traclip}

In this paper, we introduce TraCLIP as not only a trajectory-based classification approach but also a promising new direction for classification research. In this section, we conduct a series of experiments on TraCLIP and provide an in-depth analysis of its strengths, weaknesses, and limitations.

\vspace{0.3em}
\noindent
\textbf{Analysis on feature fusion.} In TraCLIP, we introduce the TFA strategy to integrate trajectory visual features into classification. Although similar to video retrieval, our focus is on utilizing complementary information from different perspectives and appearances across trajectories, rather than emphasizing temporal information. In the TFA strategy, the feature fusion module is a key component that generates enhanced trajectory features. In this work, we study five types of feature fusion mechanisms, \ie, average-based fusion, attention-based fusion (using self-attention module), self-based fusion (using self-attention and mlp modules), cross-based fusion (using cross-attention), and a concatenation-based fusion (using concatenation between visual and language features). Please refer to the \textbf{supplementary material} for detailed architectures. Fig.~\ref{fig:fusion-analysis} shows the results of different fusion mechanisms on the TETA and ClsA metrics. We can see that the second self-based fusion works generally better by achieving the best TETA score (39.4$\%$ / 33.7$\%$ for base and novel classes) and ClsA score (22.1$\%$ / 5.3$\%$ for base and novel classes). Therefore, in TRACT we employ the self-based fusion mechanism.

\vspace{0.3em}
\noindent
\textbf{Analysis on running speed.} In model design, speed is crucial as it directly affects responsiveness and user experience in real-time applications. In this work, while the additional module designs inevitably introduce some reduction in speed, we believe, as shown in Tab.\ref{tab:bank-ablation} and Tab.\ref{tab:clip-ablation}, that TRACT maintains a sufficiently fast rate and achieves a strong balance between efficiency and performance.

% \noindent
% \textbf{Analysis on training data.} 

\begin{table}[!t]\scriptsize
    \centering
    \renewcommand{\arraystretch}{1}
    \tabcolsep=2.5mm
    \caption{Ablation studies of $n_{\rm clip}$. In this study, we only evaluate the running speed of the TraCLIP module.}\vspace{-1mm}
    \resizebox{0.45\textwidth}{!}{
    \begin{tabular}{c|ccccc}
        \toprule[1.2pt]
        $n_{\rm clip}$ & \textbf{TETA} & \textbf{LocA} & \textbf{AssA} & \textbf{ClsA} & \textbf{Speed}(\textit{s/seq}) $\downarrow$ \\
        \midrule[0.8pt]
        1 & 37.96 & 53.89 & 40.50 & 19.09 & 0.77 \\
        5 & 38.59 & 54.90 & 40.51 & 20.30 & 1.28 \\
        10 & 38.48 & 54.90 & 40.51 & 20.04 & 2.55 \\
        15 & 38.51 & 54.90 & 40.51 & 20.13 & 4.97 \\
        \bottomrule[1.2pt]
    \end{tabular}
    \label{tab:clip-ablation}\vspace{-5mm}
    }
\end{table}

\subsection{Ablation Study}

To further analyze TRACT, we conduct ablations on the validation set of OV-TAO with YOLO-World as the detector. 

\vspace{0.3em}
\noindent
\textbf{Ablation on three key strategies.} In this paper, we propose three key trajectory-based strategies, \ie, TCR, TFA, and TSE strategies. To assess the impact for them, we compare the performance on the validation set of OV-TAO~\cite{li2023ovtrack} using the state-of-the-art open-vocabulary detector YOLO-World~\cite{cheng2024yolo}. As depicted in Tab.~\ref{tab:strategies-ablation}, we can see that the version incorporating all three strategies achieves the best performance across almost all metrics, especially with a notable +3.4$\%$ improvement in the ClsA metric. Besides,  As shown in Tab.~\ref{tab:tse}, although the TSE module has a limited impact on overall classification performance, it improves the classification of novel classes, which is a key goal in OV-MOT. Please note that TRACT does not involve adjustments in localization, so the LocA metric shows no significant change.

\vspace{0.3em}
\noindent
\textbf{Ablation on lengths $n_{\rm bank}$ and $n_{\rm clip}$.} To investigate the impact of two key length parameters $n_{\rm bank}$ and $n_{\rm clip}$ of TRACT, we conduct experiments with varying parameter settings. $n_{\rm bank}$ is the maximum length of the feature banks and category banks used in TCR, while $n_{\rm clip}$ is the sample clip length of TraCLIP. From Tab.~\ref{tab:bank-ablation}, we can see that, when $n_{\rm bank} = 15$, the overall best performance is achieved. Please note, in the ablation study of $n_{\rm bank}$, we exclusively apply the TCR strategy to ensure a fair comparison. We measure the model speed by the average processing time per sequence (\textit{s/seq)}, finding that increasing $n_{\rm bank}$ does not result in a notable increase in time costs (see Tab.~\ref{tab:bank-ablation}). Furthermore, as shown in Tab.~\ref{tab:clip-ablation}, we do not observe a significant improvement in effectiveness as $n_{\rm clip}$ increases, instead, there is a noticeable decrease in processing speed (see Tab.~\ref{tab:clip-ablation}). Therefore, in TRACT we use $n_{\rm clip} = 5$.

\vspace{0.3em}
\noindent
\textbf{Ablation on weighting parameter $\alpha$.} We propose the TCR strategy, where we use the weighting parameter $\alpha$ to balance the use of the track memory and feature bank. Please note, in this experiment, only TCR strategy is applied. As shown in Tab.~\ref{tab:alpha-ablation}, we can observe that when $\alpha = 0.25$, the model achieves the overall best performance.

\subsection{Discussion} \label{sec:discussion}

% \begin{figure}[t]
%  \centering
%  \includegraphics[width=0.99\linewidth]{img/Fig-Discussion.pdf}
%  %\vspace{-1em}
%  \caption{A visual comparison between Conventional MOT detections (a), OV-MOT detections (b), the complete GT annotations (c), and the GT annotations provided by TAO (d). It clearly reveals two points: first, the detection density of OV-MOT is significantly higher than that of Conventional MOT, and second, the GT annotations provided by TAO are incomplete.}
%  \label{fig:discussion}
%  % \vspace{-5mm}
% \end{figure}

\noindent
\textbf{Challenge in OV-MOT.} Current OV-MOT faces severe challenges in association due to dense detection results. We find that OV-MOT task, typically evaluated with the TETA metric\cite{li2022tracking}, has a much higher detection density than conventional MOT, which uses the HOTA metric~\cite{luiten2021hota}. Please kindly refer to the \textbf{supplementary material} for visualization of this situation. This density arises from incomplete annotations in the TAO dataset~\cite{dave2020tao}, which covers over 800 categories but contains many \textit{missing} labels. The TETA metric mitigates this by not penalizing unmatchable predictions, but this reduces penalties for false positives, prompting detectors to lower thresholds to capture rare categories. This results in dense, low-quality detections, complicating association further. We argue that the primary issues of current OV-MOT lie in data and evaluation protocol, and hope the community to address these foundational challenges.

\vspace{0.3em}
\noindent
\textbf{Can trajectory improves localization?} This paper primarily investigates using trajectory information to enhance \textit{association} and \textit{classification}, but we believe it can also aid in \textit{localization}. In OVD, localizing unknown or rare classes is challenging. However, in OV-MOT, once a target is detected, its appearance can improve localization in subsequent frames. Though preliminary experimental results following MOTRv2~\cite{zhang2023motrv2} show limited improvement, we believe it is a potential area for future research.

% In this paper, we primarily explore the use of trajectory information to aid in \textit{association} and \textit{classification}. However, we argue that it can also contribute to the localization step. In the process of OVD, one of the challenges is how to localize unknown classes or rare classes. In OV-MOT, after detecting a target, we successfully obtain its appearance features, which can help in better localizing the target in subsequent video frames. We follows the approach of MOTRv2~\cite{zhang2023motrv2} for our experiments, but the preliminary results indicate that this method does not yield significant improvements. We believe this indicates a potential area for future improvement.

\begin{table}[!t]\scriptsize
    \centering
    \renewcommand{\arraystretch}{1}
    \tabcolsep=3mm
    \caption{Ablation studies for the weighting parameter $\alpha$.}\vspace{-1mm}
    \resizebox{0.35\textwidth}{!}{
    \begin{tabular}{c|cccc}
        \toprule[1.2pt]
        $\alpha$ & \textbf{TETA} & \textbf{LocA} & \textbf{AssA} & \textbf{ClsA} \\
        \midrule[0.8pt]
        0.1 & 37.41 & 54.96 & 40.39 & \textbf{17.60} \\
        0.2 & 37.49 & 54.98 & 40.17 & 17.56 \\
        0.25 & \textbf{37.62} & \textbf{55.04} & \textbf{40.58} & 17.27 \\
        0.3 & 37.50 & 55.02 & 40.45 & 16.89 \\
        0.4 & 37.17 & 54.89 & 39.56 & 16.12 \\
        \bottomrule[1.2pt]
    \end{tabular}
    \label{tab:alpha-ablation}\vspace{-5mm}
    }
\end{table}
\section{Conclusion}

% In this work, we address the limitations of existing Open-Vocabulary Multi-Object Tracking (OV-MOT) methods by integrating trajectory-level information into both \textit{association} and \textit{classification} steps. Unlike previous approaches that primarily rely on instance-level data, our method, termed TRACT, leverages features such as trajectory characteristics and temporal dynamics to tackle OV-MOT’s key challenges. Specifically, we introduce TCR strategy to enhance identity and category consistency for trajectory-enhanced association. Furthermore, we propose TraCLIP, a plug-and-play approach that uses TFA and TSE strategies to achieve trajectory-assisted classification from visual and language perspectives, respectively. Our experiments show that TRACT substantially improves tracking performance, highlighting the value of trajectory information in open-vocabulary scenarios. This approach encourages a new perspective on OV-MOT, inviting future work to view video tracking as a unified process beyond detection and association alone. By introducing TRACT, we aim to inspire further exploration of trajectory-aware methods, paving the way for more resilient and adaptable open-vocabulary tracking solutions.

In this work, we explore trajectory-level information to improve OV-MOT  by enhancing association and classification steps. Our method, TRACT, utilizes trajectory and temporal information to enhance performance compared to instance-level approaches. We introduce the TCR strategy to improve identity and category consistency in trajectory-enhanced association and propose TraCLIP, which employs TFA and TSE strategies for trajectory-assisted classification from visual and language perspectives. Our extensive experiments show that TRACT significantly enhances tracking performance, highlighting the importance of trajectory information in open-vocabulary contexts. %We hope this work will promote further exploration of trajectory-aware ov-tracking methods.

% {
%     \small
%     \bibliographystyle{ieeenat_fullname}
%     \bibliography{main}
% }


\end{document}